%% file: iclr2027_conference.tex
\documentclass{article} 
\usepackage{iclr2027_conference,times}

\input{math_commands.tex}

\usepackage{hyperref}
\usepackage{url}
\usepackage{graphicx}

\usepackage[labelformat=simple]{subcaption}

\usepackage[font=small,labelfont=bf]{caption}

\title{Encoder-Free Multimodal LLMs\\Learn Implicit Encoders}

\author{
  \textbf{David Harwath}\textsuperscript{1} \And
  \textbf{Desh Raj}\textsuperscript{2} \And
  \textbf{Xavier Velez}\textsuperscript{3} \And
  \textbf{Sandra Arcos-Holzinger}\textsuperscript{3,4} \AND
  \textbf{Willie Jeng}\textsuperscript{1} \And
  \textbf{Karen Livescu}\textsuperscript{5} \And
  \textbf{Leibny Paola Garcia-Perera}\textsuperscript{3} \And
  \textbf{Sanjeev Khudanpur}\textsuperscript{3} \AND
  \small\mdseries
  \textsuperscript{1}The University of Texas at Austin \quad
  \textsuperscript{2}NVIDIA \quad
  \textsuperscript{3}Johns Hopkins University \\
  \small\mdseries
  \textsuperscript{4}The University of Melbourne \quad
  \textsuperscript{5}Toyota Technological Institute at Chicago
}

\iclrfinalcopy 
\begin{document}

\maketitle

\begin{abstract}
We empirically analyze two recently proposed encoder-free multimodal LLMs, Gemma-4 12B and Inkling-Small. Unlike previous multimodal LLMs, these models do not use modality-specific encoders to extract tokenized embeddings representing audio and image input; instead, they directly feed windows of samples, pixels, or spectral features to the Transformer language model. We analyze whether the early layers of these models learn to function similarly to dedicated audio and visual encoders. We find that these early layers do in fact act as ``implicit encoders,'' encoding audio and visual inputs in a way that is largely invariant to their cross-modal context. Furthermore, when performing multimodal reasoning tasks, we show that independently encoding the modalities throughout these early model layers achieves the same accuracy as jointly encoding them from the first layer onwards. Finally, we use layer-wise probing to analyze the salient information captured by the implicit encoders and find their behavior to be similar to several popular encoder models.
\end{abstract}

\section{Introduction}
\label{sec:intro}
Multimodal large language models (LLMs) such as ChatGPT~\citep{chatgpt}, Claude~\citep{claude}, Gemini~\citep{gemini}, Qwen-Omni~\citep{qwen-omni} natively process text alongside general audio and visual inputs. Because this style of model is capable of handling a wide range of both multimodal and single-modality tasks, they are rapidly displacing modality- and task-specific models. Most multimodal LLMs follow the same architectural blueprint: pre-trained, modality-specific encoders are used to extract tokenized representations of visual and audio inputs. A text LLM backbone is pre-trained separately, and the two are stitched together via a connector layer. The model is then adapted through connector-only, LoRA, or full fine-tuning.

However, as LLM backbones become larger and larger, some researchers have started to investigate whether the role of the encoders can be completely subsumed by the LLM itself.
To this end, there have been several very recently introduced ``encoder-free'' multimodal LLMs which instead tokenize audio and visual inputs by applying sliding windows (possibly with additional lightweight processing, such as a linear projection) to low-level representations, such as image pixels, waveform samples, or spectral time-frequency energies. Recent examples of this style of model are Google's Gemma-4 12B~\citep{gemma4} and the Inkling model~\citep{inkling} from Thinking Machines. The fact that these models still achieve strong performance on a range of tasks calls into question the necessity of separate audio and visual encoders. However, the pros and cons of this approach are still not fully understood. Encoder-free models offer a simpler, monolithic model architecture and have the potential to perform very early fusion of cross-modal information. However, current models (such as Gemma-4 12B) require their inputs to adhere to a specific multimodal format, and they may be less efficient to train than more modular models, especially if a large portion of their training time is spent re-learning representations that are highly similar to what a separate, pre-trained encoder would have learned anyway.

In this paper, we put forth the hypothesis that the encoder-free Gemma-4 12B and Inkling-Small models in fact \textit{rediscover} audio and image encoders using their early Transformer layers. We empirically test for implicit encoder-like behavior in these models using several techniques. We first confirm the invariance of audio and image token representations across model layers as their cross-modal context is altered, and use activation patching~\citep{meng2022locating, activation_patching} to show that independently encoding the modalities throughout the Transformer layers achieves similar performance, on a multimodal reasoning task, to jointly encoding them from the first layer onwards. Next, we perform layer-wise analysis of the speech audio representations learned by the ``implicit encoder'' layers of the LLMs, and find that they are highly similar to the representations of popular encoders such as WavLM~\citep{Chen2021WavLMLS} and HuBERT~\citep{hsu2021hubert}. Finally, we test the ``implicit encoder'' layers \textit{as encoders} by evaluating them on standard encoder evaluation benchmarks such as SUPERB~\citep{superb} and MIEB~\citep{mieb}, finding that on some tasks they are competitive, but on other tasks they fall short of existing encoders that have far smaller parameter counts. The outcomes of all of these experiments support our hypothesis that Gemma-4 12B and Inkling-Small do learn audio and image encoder-like behavior within their early layers.


\section{Related Work}

\subsection{Pre-Trained Encoders and Multimodal LLMs}
The dominant paradigm for building multimodal LLMs is to attach one or more
pre-trained, modality-specific encoders to a pre-trained text LLM. Vision Transformers (ViTs)~\citep{vit} such as CLIP~\citep{clip} or
SigLIP~\citep{siglip} are often used as image encoders, while pre-trained speech recognizers such as Whisper~\citep{whisper} and USM~\citep{usm} or self-supervised models such as wav2vec 2.0~\citep{baevski2020wav2vec}, HuBERT~\citep{hsu2021hubert}, and
WavLM~\citep{Chen2021WavLMLS} are often used to encode audio. The encoder output is then mapped into the backbone LLM's token
embedding space by a ``connector'', which is often a linear projection or lightweight MLP as in LLaVA~\citep{llava}, SALMONN~\citep{salmonn}, and Qwen2-Audio~\cite{qwen2audio}.
 Multimodal models such as Qwen-Omni~\cite{qwen-omni},
MOLMo~\cite{molmo}, and Gemini~\cite{gemini} extend this framework to three or more
modalities by using parallel encoders for audio and visual inputs, alongside text tokenizers typically based on byte-pair encoding~\citep{bpe} and a text token embedding table.

\subsection{Encoder-Free Multimodal LLMs}
A recently introduced type of multimodal LLM is the ``encoder-free'' architecture, in which raw pixel patches or windows of low-level audio features (waveform samples or spectral features) are projected into the input token space of a text LLM. The major motivations for this approach are twofold: first, it offers a simpler, monolithic model rather than an ensemble of pre-trained encoders assembled to a backbone LLM. Second, it allows the encoders to be jointly optimized with the LLM backbone, and potentially allows for richer fusion of cross-modal information from the earliest layers of the model. Early examples of this paradigm applied to Vision-and-Language models (VLMs) are Fuyu-8B~\citep{fuyu}, EVE~\citep{eve,eve2}, and SOLO~\citep{solo}. Gemma-4 12B~\citep{gemma4} and Inkling~\citep{inkling} are the first examples of large-scale, open-weight, encoder-free LLMs that simultaneously handle text, audio, and image input. They form the basis of our studies in this paper, and so we discuss their architectures in detail in the following two paragraphs.

\textbf{Gemma-4 12B} is an open-weight, 48 layer, decoder-only Transformer LLM. It is capable of processing text, image, video, and audio inputs with a unique encoder-free architecture. While the other models in the Gemma-4 family utilize large, pre-trained ViT~\citep{vit} vision encoders and USM~\citep{usm} audio encoders, the Gemma-4 12B model was trained from scratch using only linear projections to embed image patches and waveform windows into its token embedding space. Audio waveform samples at a 16kHz sampling rate are captured with non-overlapping sliding windows of 640 samples (40ms) before being projected into the 3840-dimensional input token space with a single linear transform. Similarly, non-overlapping RGB image pixel patches of size 48x48x3 are projected into the same 3840-dimensional token embedding space before being concatenated with the text and audio inputs to the model. Finally, an important detail about Gemma-4 12B that pertains to our analysis is the fact that it uses interleaved global and sliding window attention across its Transformer layers at a 5:1 ratio. For all of our experiments, we use the instruction-tuned version of Gemma-4 12B.

\textbf{Inkling} is a recently-released, open-weight, multimodal LLM capable of processing text, image, and audio inputs. We conduct our experiments on the Inkling-Small model, a 42-layer, decoder-only, Tranformer LLM that uses a sparse Mixture of Experts~\citep{moe} architecture. The model has 276B total and 12B active parameters, and was trained on 45T total tokens spanning text, audio, image, and video. Similarly to Gemma-4 12B, Inkling-Small interleaves global and sliding-window self-attention with a 5:1 ratio, and is also encoder-free, although its architecture differs from that of Gemma-4 12B. 40x40x3 patches of RGB image pixels are encoded with a 4-layer MLP into 4096-dimensional tokens. Audio frames are represented as dMel~\citep{dmel} quantized spectral features, which are mapped into the 4096-dimensional tokens via a linear projection.

\subsection{Representation Analysis and Emergent Modularity}
Many prior works have studied the representations that emerge within deep neural network models. A recurring theme throughout this work is the emergence of a depth-ordered hierarchy of abstraction. For example, \citet{zeiler2014visualizing} showed that successive convolutional layers progress from representations of oriented edges through textures and object parts to whole-object
detectors, and \citet{pasad2021layer,pasad2023comparative,pasad2024words} made similar observations about how self-supervised speech audio models capture low-level acoustic features to phones and words with increasing layer depth. In the text domain, The ``BERTology''
literature~\citep{rogers2020primer,tenney2019bert} found that the BERT~\citep{devlin2018bert} model ``rediscovers'' part of-speech tagging, parsing, named-entity recognition, semantic roles, and co-reference resolution at different layers in the model.

Recent work has carried these tools to LLMs and to multimodal LLMs. \citet{stages_of_inference} propose that LLMs share
the same general stages of inference, while~\citet{wendler2024llamas} finds that multilingual models pivot through an English-like
concept space in their middle layers. Causal methods such as activation
patching~\citep{activation_patching} test whether a representation is actually used downstream. Applied to multimodal LLMs, these methods have shown that visual token
representations become progressively more language-aligned with
depth~\citep{neo2024interpreting}, and that cross-modal integration concentrates in
particular layers and token positions~\citep{crossmodal_flow}. Our work extends some of these analysis techniques to encoder-free, multimodal LLMs and aims to uncover whether parts of these models exhibit emergent specialization that fulfills the same role as the modality encoders traditionally employed in multimodal LLMs.


\section{Analysis of Gemma-4 12B and Inkling-Small}

\subsection{Context invariance of embeddings}
\label{sec:cka}

Because Gemma-4 12B and Inkling-Small take as input a concatenated sequence of text, image, and audio tokens, these inputs have the opportunity to perform cross-modal contextualization starting at the very first layer of the model. If this contextualization does occur, then the same audio input encoded multiple times with different text or image contexts will result in different representations of the audio tokens. However, our hypothesis is that the early layers of these models are acting as modality-specific encoders, and therefore will not contextualize across modalities until a sufficiently abstract representation is computed. If this hypothesis is true, then the implication is that there exists some layer $L$ such that the representation of an audio (or visual) input to the model will be invariant to changes in its cross-modal context up until layer $L$, after which it will start to diverge with varying context. We experimentally investigated the degree to which Gemma-4 12B and Inkling-Small obey this context invariance, using either audio or visual input as the anchor modality.

\subsubsection{Context Invariance of Audio Inputs}

We randomly sample 1,000 utterances from the dev-clean split of Librispeech~\citep{panayotov2015librispeech}, and form a neutral context for each of these by pairing each waveform with a neutral text string (unrelated to the speech waveform; the same string is used for all waveforms) and a black-to-white gradient image. We then encode each waveform jointly with the text and image inputs using both models, and extract their layer-wise outputs (3840-dimensional outputs across 48 layers for Gemma-4 12B, and 4096-dimensional outputs across 42 layers for Inkling-Small). These embeddings serve as the neutral reference representations.

\begin{figure}[!htbp]
  \centering
  \begin{subfigure}[t]{0.48\linewidth}
    \centering
    \includegraphics[width=\linewidth]{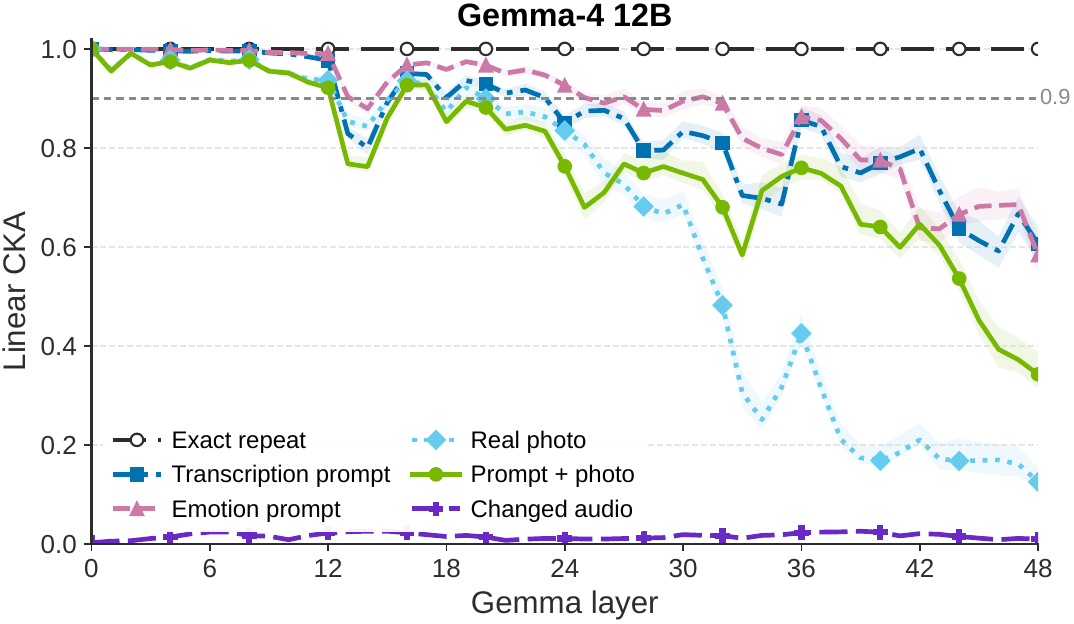}
    \label{fig:gemma4_audio_cka}
  \end{subfigure}
  \hfill
  \begin{subfigure}[t]{0.48\linewidth}
    \centering
    \includegraphics[width=\linewidth]{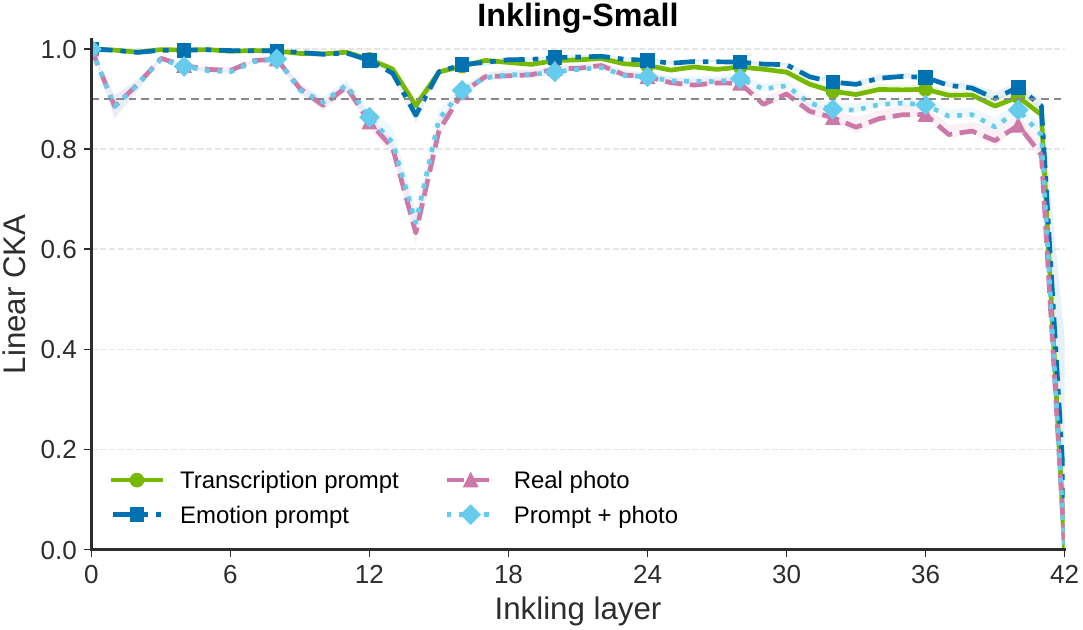}
    \label{fig:inkling_audio_cka}
  \end{subfigure}
  \caption{Layer-wise CKA similarity of audio tokens encoded under varying image and text contexts vs. audio tokens encoded with a neutral context.}
    \label{fig:audio_cka}
\end{figure}

Next, we replace the gradient image and/or neutral text with different inputs: \textbf{Transcription Prompt}: A text prompt instructing the model to transcribe the content of the input speech waveform into a word sequence. \textbf{Emotion prompt}: A text prompt instructing the model to recognize the emotional content of the input speech waveform. \textbf{Real photo}: A natural photograph, sampled from the MSCOCO~\citep{lin2014microsoft} dataset. We re-encode each speech waveform 4 times with each of the models: once for each of the three context changes listed above, and a 4th time that simultaneously replaces the gradient image with a natural photo and the neutral text with the transcription prompt. Next, we use Centered Kernel Analysis (CKA)~\citep{pmlr-v97-kornblith19a} to compute the layer-to-layer similarity between the neutral reference representations of the audio input tokens, vs.~the same audio inputs with varying text and image contexts. The results of this analysis are depicted in Figure~\ref{fig:audio_cka}.

For Gemma-4 12B (Figure~\ref{fig:audio_cka}), across all of the context changes the CKA similarity of the audio token representations to their neutral reference remains high (well above 0.9) until layer 12, after which they gradually drop with increasing depth into the model. This suggests that the representations of the audio embeddings computed by the first 12 layers of Gemma-4 12B do not significantly change as a function of varying their textual and visual context. We would observe similar behavior if the model instead used a completely separate audio encoding trunk unable to attend to the visual or text inputs. The bulk of the drop in CKA similarity to the reference embeddings happens gradually between layers 12 and 48, indicating that cross-modal contextualization takes place throughout the latter 75\% of the model layers, presumably only \textit{after} the audio token representations have reached a sufficiently high level of abstraction. We analyze this in more depth in Section~\ref{sec:layerwise_analysis}.

Inkling-Small (Figure~\ref{fig:audio_cka}) shows an even more striking trend: aside from a small dip at layer 14 and a significant drop at the very last layer, the CKA similarity between the altered-context audio representations and the neutral-context representations remains very high across all layers of the model. This suggests that Inkling-Small does not contextualize the audio token representations with cross-modal information to a significant degree until the very last layer of the model. We analyze exactly where Inkling-Small performs cross-modal contextualization later in Section~\ref{sec:joint_encoding}.

\subsubsection{Context Invariance of Visual Inputs}
Next, we conduct a similar experiment but use the visual inputs as the anchor modality. Gemma-4 12B's canonical token layout places text tokens first, followed by image tokens, and finally followed by audio tokens. Because the model uses causally-masked self-attention, the image tokens do not have an opportunity to attend to the audio tokens. We therefore restrict our analysis of the context invariance of image embeddings to comparisons between different textual contexts. First, we randomly sample 797 images from MSCOCO~\citep{lin2014microsoft}, and pair each one with three different text contexts: \textbf{No caption}: Blank whitespace is provided to the model instead of a caption. \textbf{Matched caption}: A ground-truth matching caption for each image from MSCOCO is sampled and provided as input to the model along with its paired image. \textbf{Mismatched caption}: A randomly sampled caption belonging to a different MSCOCO image is used as the text context.

\begin{figure}[!htbp]
  \centering
  \begin{subfigure}[t]{0.48\linewidth}
    \centering
    \includegraphics[width=\linewidth]{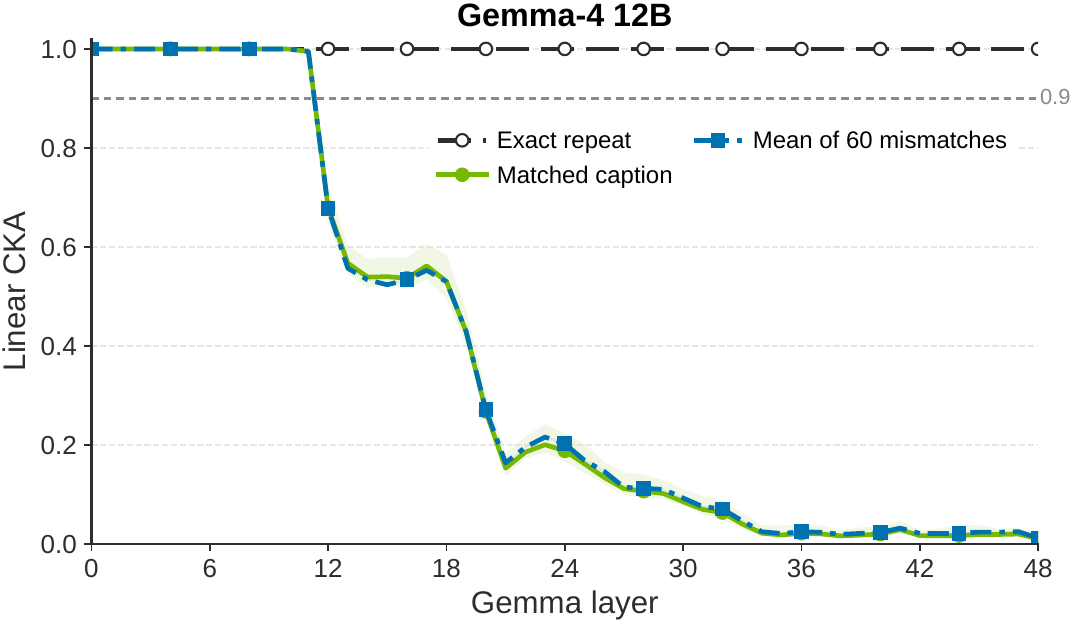}
    \caption{Analysis of image context sensitivity for the Gemma-4 12B Model.}
    \label{fig:gemma4_image_cka}
  \end{subfigure}
  \hfill
  \begin{subfigure}[t]{0.48\linewidth}
    \centering
    \includegraphics[width=\linewidth]{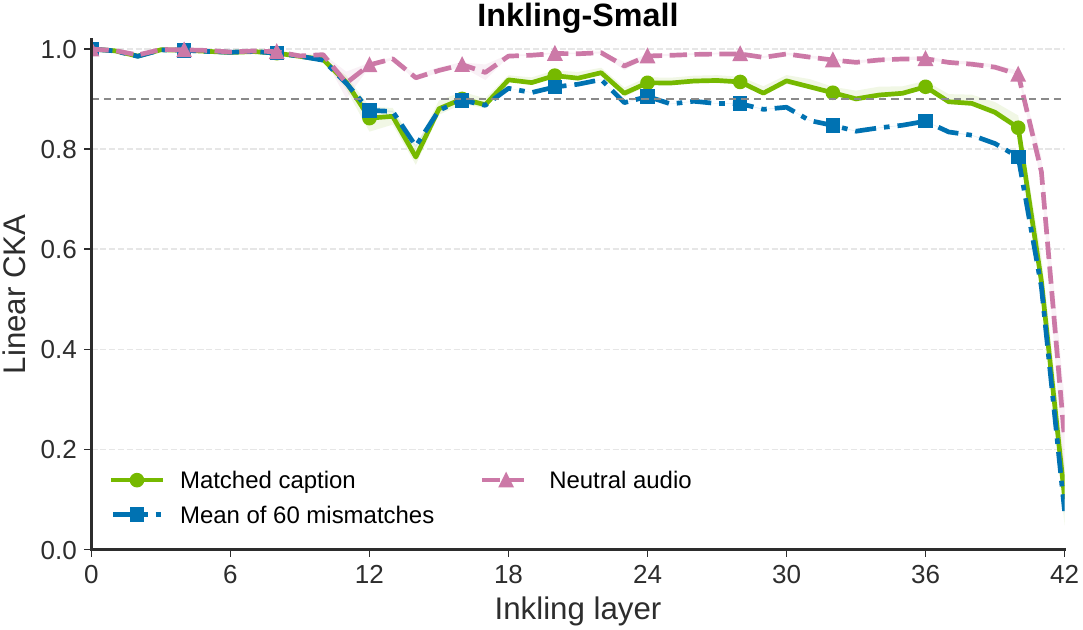}
    \caption{Analysis of image context sensitivity for the Inkling-Small Model.}
    \label{fig:inkling_image_cka}
  \end{subfigure}
  \caption{Layer-wise CKA similarity of image tokens encoded under varying text and audio contexts vs. image tokens encoded with a neutral context.}
  \label{fig:image_cka}
\end{figure}

We jointly encode each image with the ``No caption'' contextual condition and extract the per-layer outputs from Gemma-4 12B to form the reference embeddings. Next, we encode each image with its matched caption. Finally, we encode each image 60 times with 60 different mismatched captions, and average the embeddings. We then compute the CKA similarity between the reference embeddings and the embeddings computed in the context of a matched caption, and finally repeat this procedure between the reference embeddings and mismatched caption embeddings. The results of this experiment are depicted in Figure~\ref{fig:image_cka}.

In the case of Gemma-4 12B (Figure~\ref{fig:gemma4_image_cka}), we again observe similar curves to what we saw when analyzing the encoding of audio tokens with the same model (Figure~\ref{fig:audio_cka}): the CKA similarity stays extremely high during the early layers of the model, staying close to 1.0 until layer 12, at which point it rapidly drops. This provides additional evidence that the early layers of Gemma-4 12B are encoding audio and visual inputs into higher-level representations \textit{before} allowing for cross-modal contextualization, which only begins at the middle layers of the model. Inkling-Small (Figure~\ref{fig:inkling_image_cka}) again behaves differently from Gemma-4 12B, with its image patch embeddings staying almost completely invariant to cross-modal context until layer 10, at which point they become slightly sensitive to context and remain so until layer 42. This behavior is very similar to what we observed when examining how the same model encodes audio tokens (Figure~\ref{fig:audio_cka}). 

\subsection{Independent vs.~Joint Encoding of Modalities for Cross-Modal Reasoning}
\label{sec:joint_encoding}

Our next experiment follows up on our context-invariance analysis by tasking the models with a question answering task that requires simultaneous reasoning across text, image, and audio inputs. Nominally, both Gemma-4 12B and Inkling-Small would take all three modalities as input simultaneously and perform joint encoding. In addition to this condition, we perform a second forward pass that uses a form of activation patching~\citep{activation_patching}: We separately encode the tokens belonging to each modality (isolated from all other modalities, with one forward pass per modality) up until layer $L$, extract these neutral-context representations at layer $L$ for all modalities, concatenate them together according to the model's expected token layout, and then feed these concatenated tokens into the remainder of the model beginning at layer $L+1$. Informed by our analysis in Section~\ref{sec:cka}, we hypothesize that if the inputs to the models are invariant to changes in their cross-modal context up until some layer $L$, then there should be no difference in their performance on the question answering task using jointly vs. independently encoded inputs up until layer $L$; after layer $L$, we should expect to see changes in the performance of the model. 

To implement this experiment, we use the OmniBench~\citep{omnibench} dataset which is comprised of 1,142 quadruplets, each containing a text question, a visual image, an audio waveform, and a text answer. The evaluation examples in OmniBench were constructed in such a way that each question should only be answerable by taking into account both the image and the audio waveform. For each of the 1,142 examples in OmniBench, we separately encode the question text, image, and audio waveform with three separate forward passes through the model. When encoding one of these modalities, we preserve its positional information by filling in the tokens belonging to the other modalities with padding. This results in the token embeddings retaining their positional embeddings, and also keeps the number of tokens in their surrounding context constant.

\begin{figure}
  \begin{minipage}[b]{0.54\linewidth}
    \centering
    \includegraphics[width=\linewidth]{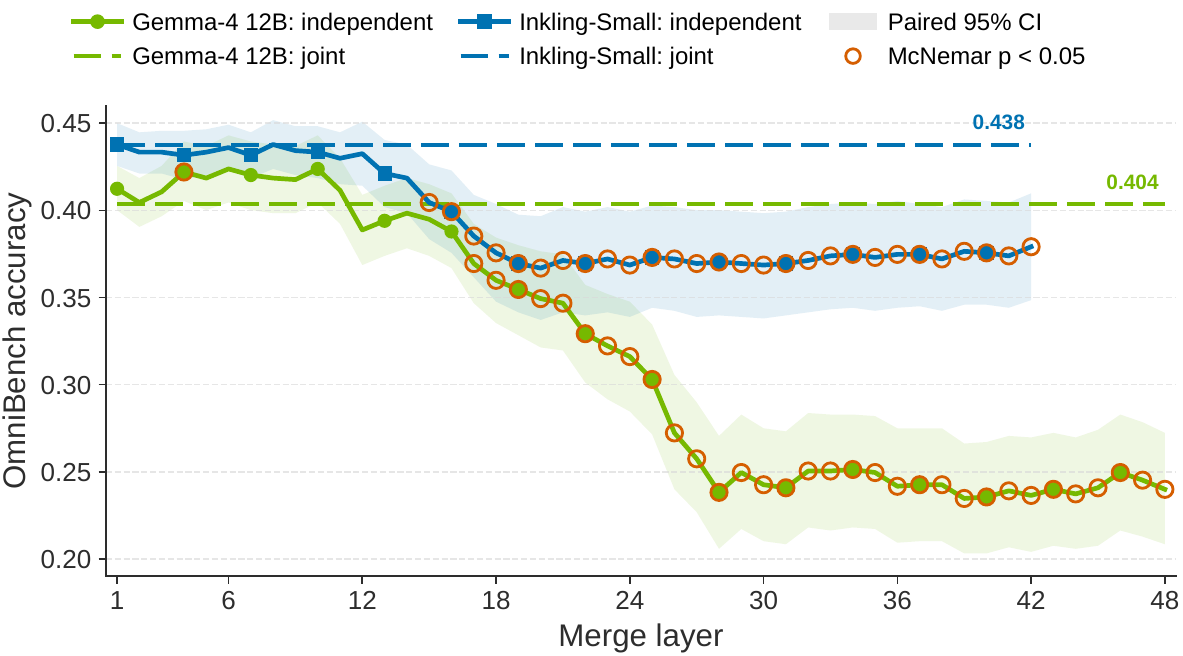}
    \caption{Performance of Gemma-4 12B and Inkling-Small on OmniBench as a function of the model layer where independently-encoded modality representations are merged.}
    \label{fig:gemma4_inkling_omnibench}
  \end{minipage}
  \hspace{0.2cm}
  \begin{minipage}[b]{0.46\linewidth}
    \centering
    \includegraphics[width=\linewidth]{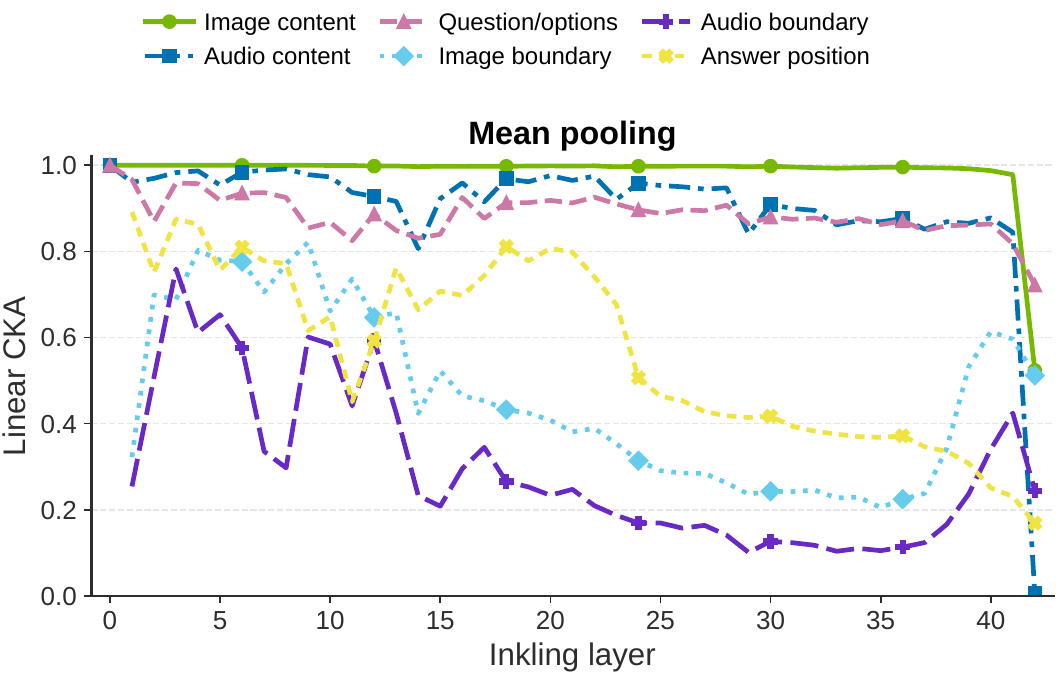}
    \caption{Similarity of jointly versus independently encoded representations for different token segments: content tokens (image, audio, text) and boundary tokens.}
    \label{fig:inkling_segmentwise_cka}
  \end{minipage}
\end{figure}

The results of our OmniBench analysis on both models is shown in Figure~\ref{fig:gemma4_inkling_omnibench}. The blue dashed line indicates the 43.8\% baseline accuracy that Inkling-Small achieves on the task when all modalities are jointly encoded, and the green dashed line indicates the 40.4\% baseline accuracy achieved by Gemma-4 12B under the same condition. We observe that the performance of both models stays relatively constant around their baseline performance until Layer 12, at which point it begins to drop---degrading to chance level (25\%) in the case of Gemma-4 12B, and dropping to approximately 37\% for Inkling-Small. On the surface, these results are consistent with our observations of Gemma-4 12B in Section~\ref{sec:cka}. We saw significantly higher context sensitivity in the model's later layers compared to Inkling-Small, and here we see a correspondingly large drop in OmniBench accuracy when merging activations at these later layers.

Conversely, our analysis in the previous section suggested that cross-modal context has very small impact on Inkling-Small's audio embeddings (Fig.~\ref{fig:audio_cka}) and image embeddings (Fig.~\ref{fig:inkling_image_cka}). This seems contrary to its performance on OmniBench, where we see a marked degradation in the independently-embedded modalities after Layer 12 (although not as severe as Gemma-4). To explain this discrepancy, we computed the linear CKA similarity scores on Inkling's segment-wise embeddings between the jointly and independently embedded scenarios, as shown in Fig.~\ref{fig:inkling_segmentwise_cka}. Here, a segment refers to either \textit{content} tokens (audio, image, question) or \textit{boundary} tokens. While the similarity scores for the content segments remain high throughout the layers, the boundary tokens see a sharp decline after Layer 12. This suggests that Inkling-Small stores its cross-modal context primarily in the boundary tokens between the modality token segments. 

\begin{table*}[t]
    \centering
    \caption{Performance of the first 12 layers of Gemma-4 12B and Inkling-Small compared to baseline SSL models on SUPERB. We also report the performance of our distilled Gemma models (Section~\ref{sec:distillation})}
    \label{tab:superb_results}
    \renewcommand{\arraystretch}{1.2}
    {\scriptsize
    \begin{tabular}{l|cccc|ccccccc}
        \hline
        \hline
        & \multicolumn{4}{c|}{\textbf{Mean-Pooled Embeddings}} & \multicolumn{7}{c}{\textbf{Frame-wise Embeddings}} \\
        \cline{2-12}
        \textbf{Model} & KS & IC & SID & ER & PR & ASR & QbE & \multicolumn{2}{c}{SF} & ASV & SD \\
        & Acc$\uparrow$ & Acc$\uparrow$ & Acc$\uparrow$ & Acc$\uparrow$ & PER$\downarrow$ & WER$\downarrow$ & MTWV$\uparrow$ & F1$\uparrow$ & CER$\downarrow$ & EER$\downarrow$ & DER$\downarrow$ \\
        \hline
        FBANK & 8.63 & 9.10 & 8.5E-4 & 35.39 & 82.01 & 23.18 & 0.0058 & 69.64 & 52.94  & 9.56  & 10.05 \\
        \hline
        wav2vec 2.0 Base & 96.23 & 92.35 & 75.18 & 63.43 & 5.74 & 6.43 & 0.023 & 88.30 & 24.77 & 6.02 & 6.08 \\
        wav2vec 2.0 Large & 96.66 & 95.28 & 86.14 & 65.64 & 4.75 & 3.75 & 0.049 & 87.11 & 27.31 & 5.65 & 5.62 \\
        HuBERT Base & 96.30 & 98.34 & 81.42 & 64.92 & 5.41 & 6.42 & 0.074 & 88.53 & 25.20 & 5.11 & 5.88 \\
        HuBERT Large & 95.29 & 98.76 & 90.33 & 67.62 & 3.53 & 3.62 & 0.035 & 89.81 & 21.76 & 5.98 & 5.75 \\
        \hline
        Gemma 4 12B & 96.79 & 98.13 & 85.14 & 66.79 & 11.11 & 9.57 & 0.003 & 85.67 & 30.13 & 6.68 & 7.85 \\
        Inkling-Small & 94.90 & 98.29 & 85.52 & 69.61 & 15.44 & 19.12 & 0.006 & 71.94 & 50.25 & 6.96 & 12.14 \\
        \hline
        Gemma Distilled 800M & 96.69 & 98.26 & 92.44 & 66.24 & 9.81 & 10.20 & 0.019 & 87.38 & 28.09 & 5.87 & 7.12 \\
        Gemma Distilled 400M & 96.62 & 97.63 & 90.55 & 65.55 & 9.85 & 11.08 & 0.016 & 85.65 & 30.41 & 6.19 & 7.54 \\
        Gemma Distilled 200M & 97.18 & 95.25 & 86.26 & 66.93 & 10.38 & 12.02 & 0.018 & 86.63 & 28.68 & 6.20  & 7.89 \\
        Gemma Distilled 100M & 96.92 & 96.73 & 81.67 & 65.40 & 11.31 & 13.03 & 0.016 & 85.51 & 31.16 & 6.77 & 9.92 \\
        \hline
        \hline
    \end{tabular}%
    }
    \vspace{1mm}
    
    \tiny{KS: Keyword Spotting, IC: Intent Classification, SID: Speaker Identification, ER: Emotion Recognition, PR: Phoneme Recognition, ASR: Automatic Speech Recognition w/o LM, SF: Slot Filling, ASV: Automatic Speaker Verification, SD: Speaker Diarization.}
\end{table*}

\subsection{Performance As An Audio Encoder} Although our previous experiments have provided evidence that the early layers of both Gemma-4 12B and Inkling-Small do behave as image and audio encoders, the question arises as to whether these learned representations can generalize across modality-specific tasks. 
To that end, we evaluate how well the frame-level audio embeddings computed by both models perform in various downstream speech processing tasks using SUPERB~\citep{superb}, which is a widely-used evaluation benchmark for assessing the performance of pre-trained speech embedding models across a range of tasks including speech recognition, emotion recognition, speaker identification, and more.
We compare our performance with the numbers reported for both the Base and Large variants of the HuBERT~\citep{hsu2021hubert} and wav2vec 2.0~\citep{baevski2020wav2vec} SSL models on SUPERB. 
We also include the reported log mel filterbank features (FBANK) as an additional baseline since they are a popular feature used in many speech tasks. 

Table \ref{tab:superb_results} separates the tasks into two categories: those that use a single mean-pooled embedding per layer and those that use all of the frame-wise embeddings per-layer. 
The embeddings of the first 12-layers are then combined using a weighted-sum to get the final output. 
The tasks that use mean-pooled embeddings are all classification tasks and measure accuracy. 
We see that for both encoder-free models, the performance on these tasks is competitive with the established baselines. 
This demonstrates that the early layers of both models function as audio encoders for purposes of these tasks, producing global features that generalize across different speech classification tasks.

On the tasks that require more granular representations, such as phoneme and speech recognition, we surprisingly see that the frame-wise embeddings under-perform relative to the baseline SSL models on tasks such as ASR and phone recognition. This is consistent with the ASR results reported by~\citet{gemma4_asr_benchmark}, which measured Gemma-4 12B's ASR performance to be substantially worse than the Gemma-4 2B and 4B models - for example, the 12B model reaches a WER of 23.9\% on GigaSpeech, while the 2B and 4B models respectively reach 11.8\% and 11.1\% on the same task.


\begin{table*}[t]
    \centering
    \caption{Frozen layer-10 image representations of Gemma-4 12B compared to dedicated vision encoders on the MIEB-Lite image-only tasks.
    Classification reports accuracy (Acc) and clustering reports Normalized Mutual Information (NMI).
    Best per task in \textbf{bold}.}
    \label{tab:mieb_lite}
    \renewcommand{\arraystretch}{1.2}
    {\scriptsize
    \begin{tabular}{l|cccccccc|cc|c}
        \hline
        \hline
        & \multicolumn{8}{c|}{\textbf{Image Classification} (Acc $\uparrow$)} & \multicolumn{2}{c|}{\textbf{Clustering} (NMI $\uparrow$)} & \\
        \cline{2-11}
        \noalign{\vskip 3pt}
        \textbf{Model}
        & \rotatebox{90}{Country211}
        & \rotatebox{90}{DTD}
        & \rotatebox{90}{EuroSAT}
        & \rotatebox{90}{GTSRB}
        & \rotatebox{90}{OxfordPets}
        & \rotatebox{90}{PatchCamelyon}
        & \rotatebox{90}{RESISC45}
        & \rotatebox{90}{SUN397}
        & \rotatebox{90}{ImageNetDog15}
        & \rotatebox{90}{TinyImageNet}
        & \textbf{Avg} \\
        \hline
        SigLIP2-base & \textbf{18.05} & \textbf{79.48} & \textbf{86.40} & \textbf{80.76} & 92.92 & \textbf{73.52} & \textbf{88.36} & \textbf{77.16} & 79.61 & 59.82 & \textbf{73.61} \\
        DINOv2-base & 10.73 & 75.93 & 81.66 & 53.48 & \textbf{94.12} & 72.73 & 81.30 & 71.03 & \textbf{85.67} & \textbf{74.31} & 70.10 \\
        \hline
        Gemma-4 12B (L10) & 10.10 & 71.90 & 83.20 & 48.10 & 53.70 & 72.70 & 78.10 & 56.40 & 50.40 & 59.30 & 58.39 \\
        \hline
        \hline
    \end{tabular}%
    }
    \vspace{1mm}
\end{table*}

\subsection{Performance as a vision encoder}

Similarly, we evaluate whether the representation learned in these early layers for both Gemma-4 12B and Inkling-Small behave as a vision encoder.
We use the image-only tasks found in the lite subset of the MIEB~\cite{mieb} benchmark.
This gives us 10 tasks total, split into 8 classification and 2 clustering tasks.
For Gemma-4 12B, we probe over the first 12 layers, as shown in Figure~\ref{fig:gemma4_image_classification} and Figure~\ref{fig:gemma4_image_clustering} (See Appendix Sec.~\ref{sec:appendix_mieb}), and find that the representation provided by the 10th layer gives the highest overall score on the given tasks.
Table~\ref{tab:mieb_lite} shows that performance on clustering underperforms compared to the two vision encoders.
 In classification tasks, it is competitive on half the tasks, while it is underperforming on the others.

\subsection{Layer-wise analysis of audio representations}
\label{sec:layerwise_analysis}

\begin{figure}
    \centering
    \includegraphics[width=\linewidth]{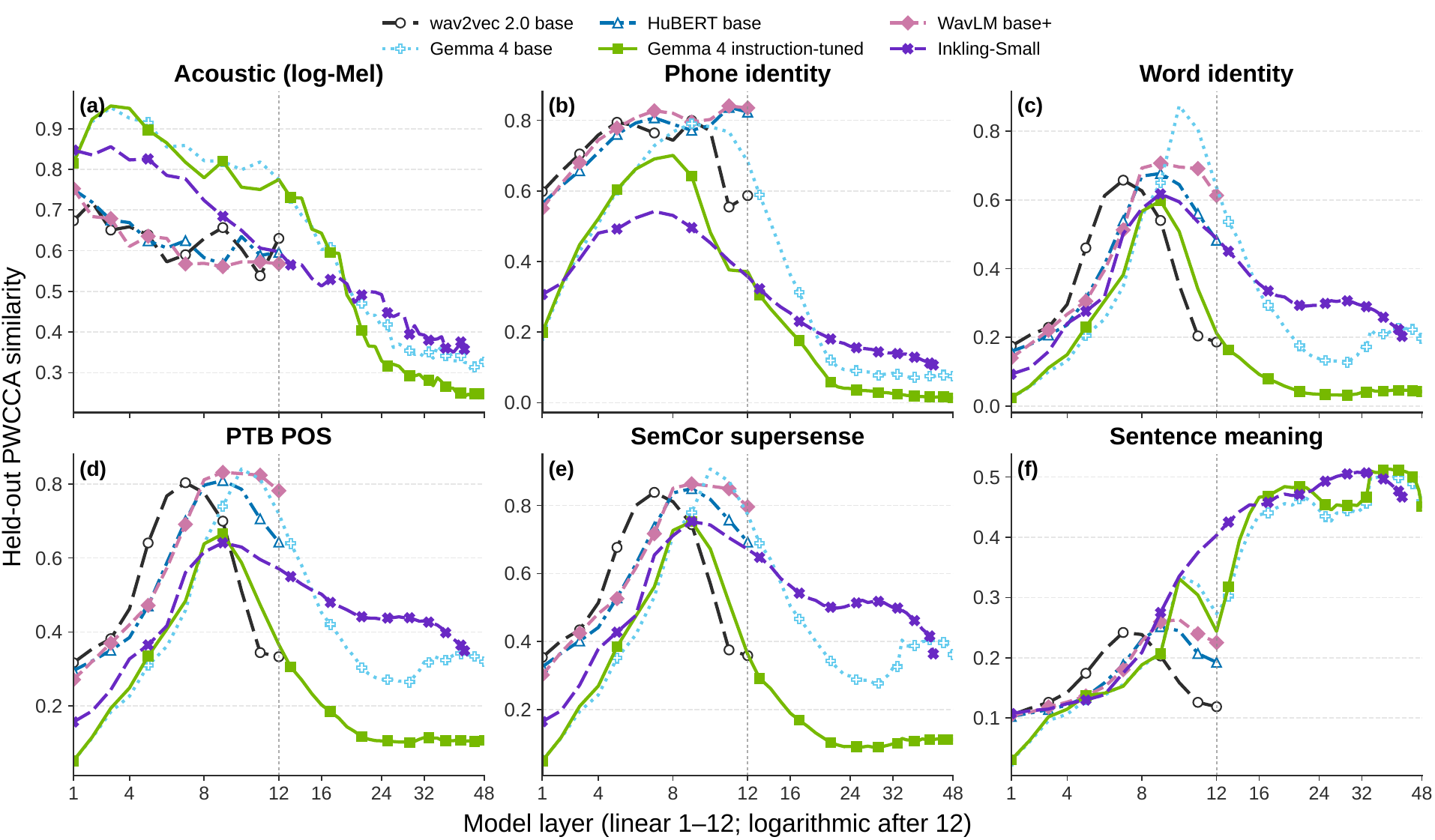}
    \caption{Layer-wise analysis of the audio token representations learned by Gemma-4 12B, Inkling-Small, and several self-supervised baselines.}
    \label{fig:layerwise_analysis}
\end{figure}

Our next analysis treats the early layers of Gemma-4 12B and Inkling-Small as audio encoders, and probes what kind of information is represented at each layer. We closely follow the work of \citet{pasad2021layer,pasad2023comparative}, who performed layer-wise analysis of the information (acoustic, phonetic, lexical, semantic, etc.) present in the different layers of self-supervised speech encoders such as HuBERT~\citep{hsu2021hubert}, WavLM~\citep{Chen2021WavLMLS}, and wav2vec2.0~\citep{baevski2020wav2vec}. In~\citet{pasad2021layer,pasad2023comparative}, the authors show that while these models do exhibit meaningful differences from one another, a common pattern in their representations does emerge: the lower levels of these models tend to capture low-level acoustic information, the middle to later layers capture phonetic information, and the layers after that capture information about word identity and meaning. 

We replicated the analysis pipeline of~\citep{pasad2021layer,pasad2023comparative} and applied it to both Gemma-4 12B and Inkling-Small. Specifically, we used projection-weighted canonical correlation analysis (PWCCA)~\citep{pwcca} to quantify the similarity between the layer-wise representations of each model and acoustic features (via log-mel featurebank features), phones (via one-hot phonetic labels aligned at the frame level), words (via one-hot word labels aligned at the frame level), syntax (via Penn Treebank~\citep{penn_treebank} part-of-speech tag labels aligned with words), semantics (via SemCor~\citep{semcor} word-level supersenses). We also add an additional comparison to whole-sentence meaning via sentence embeddings computed using the MPNet text embedding model~\citep{song2020mpnet} applied to the text transcripts of each utterance. 

Figure~\ref{fig:layerwise_analysis} shows the results of this analysis on Gemma-4 12B, Inkling-Small, HuBERT-Base, WavLM-Base+, and wav2vec2.0-Base. Regarding the similarity to log-mel filterbank features, we observe that Gemma-4 12B starts with a very low score at layer 0 (representing the raw waveform samples) but rapidly peaks at layer 3 with a PWCCA similarity score of 0.95. This indicates that although Gemma-4 12B takes as input windows of raw waveform sample amplitudes, the first 3 layers of the model learn a transformation of those features that nearly perfectly captures log-mel filterbank energies (although this does not rule out the possibility that the layers encode additional information beyond that). It should be noted that the layer 0 scores for the HuBERT, WavLM, and wav2vec2.0 models with the log-mel filterbank energies already starts at a high value, which we attribute to the fact that these models all use several convolutional layers as a waveform encoder, and the layer 0 features represent the output of this encoder. Inkling-Small takes as input dMel features, so it is not surprising that its initial layers have a high similarity to log-Mel filterbank energies.

Regarding the similarity of Gemma-4 12B and Inkling-Small features to phonetic and word identity, we observe that phonetic similarity peaks at layer 8 for Gemma and layer 7 for Inkling, and lexical similarity peaks in both models at layer 9. Both of these patterns are similar to what is observed in the self-supervised baselines, except for the fact that the phonetic information in the self-supervised models stays relatively high after its initial peak. Conversely, both phonetic and word information in Gemma-4 12B and Inkling-Small drop rapidly after their peaks, raising the question of what information is being captured by the models at layer 10 and onward.

In the bottom of Figure~\ref{fig:layerwise_analysis}, we try to answer this question by measuring the similarity of Gemma-4 12B's and Inkling-Small's representations to higher levels of abstraction, starting with part of speech tags and word meanings (left and center plots). However, we find that both of these similarities peak alongside word identity, before rapidly falling. However, when we examine the similarity to whole-sentence embeddings we see this information begin to steadily grow at layer 10, and remain high throughout the remainder of the model. Overall, our analysis suggests that both Gemma-4 12B and Inkling-Small evolve their speech audio representations by first capturing low-level acoustic information, mapping this to phonetic sequences followed by word sequences, and then finally mapping these word sequences to representations that capture higher-level sentence meaning.

\subsection{Isolating Implicit Encoders via Distillation}
\label{sec:distillation}

Throughout the previous several sections we have shown that the early layers of Gemma-4 12B and Inkling-Small behave like encoders that process each modality independently. However, in both of these models the same set of parameters serves every modality. We ask whether the encoding functions of each modality are separable into dedicated models, and how much of the model's capacity a single modality actually requires. We focused on the audio modality, and used distillation to train a dedicated compact audio encoder from the first 12 layers of the pretrained Gemma-4 12B model. We trained randomly initialized student models of various sizes, ranging from 800M down to 100M parameters. Every student model maintained a similar architecture to that of the teacher model, preserving the 12 blocks, the query head and key/value head count, and the same attention pattern, including the alternation sliding-window at the sixth and twelfth layers. We resized the student models by scaling down the model dimension, as well as the dimension of the attention heads.


Our distillation process used approximately 8000 hours of audio sampled from GigaSpeech~\citep{gigaspeech}, Common Voice English~\citep{commonvoice}, and AudioSet~\citep{audioset}. We performed layer-to-layer distillation using the DistilHuBERT objective~\citep{distillhubert}. 
We evaluated the quality of the distilled audio encoders on the SUPERB benchmark, the results of which can be found in Table~\ref{tab:superb_results}. Distillation largely preserves the teacher's performance even at a student model roughly 7\% of the teacher's size, suggesting that the intrinsic capacity of the implicit audio encoder in Gemma-4 12B is much less than the 2.73B parameters of the first 12 layers.


\section{Discussion}
In this paper, we analyzed the behavior of Gemma-4 12B and Inkling-Small, two recently released encoder-free multimodal LLMs. We hypothesized that the early layers of these models would learn to function similarly to traditional modality-specific encoders found in other multimodal LLMs, and conducted a series of experiments whose results overall supported this hypothesis. We tested these implicit encoders the SUPERB and MIEB benchmarks, and found that on some tasks they are competitive with existing encoders, but achieve underwhelming performance on other tasks. We also analyzed the layer-wise representations both LLMs when used as a speech audio encoder, and found several similarities to existing speech encoders. Finally, we attempted to isolate the implicit speech audio encoder formed by the first 12 layers of Gemma-4 12B via distillation and found that we could retain most of its performance using as little as 7\% of its parameters. 

Encoder-free multimodal LLMs are appealing to researchers and  developers because they offer a simplified, monolithic model architecture and training recipe that does not need to be assembled from multiple, separately-trained components. However, this could potentially come with costs. As our analysis has shown, these models are very likely to ``re-invent the wheel'' each time they are trained from scratch, by spending model parameters and training time re-learning much of the same functionality that a pre-trained encoder could offer off-the-shelf. It is not yet clear whether from a task performance perspective encoder-free models have other inherent advantages over multimodal LLMs built with dedicated encoders. Future work along this direction should try to further quantify what advantages and disadvantages encoder-free architectures have over using separate encoders and LLM backbones in terms of task performance and training requirements (both data amount and compute time). We conjecture that it may be possible to achieve the best of both worlds, for example by initializing the early layers of an encoder-free architecture with parameters from pre-trained encoder models, or by using knowledge distillation from pre-trained encoders into the early layers of an encoder-free LLM as an additional training objective to speed up learning. 


\subsection*{AI use statement}

In this work, we used generative AI tools for assisting in experimental implementation, drafting figures, literature review and searching for related work, and editing/polishing writing.
We have not used generative AI tools for any other tasks with required disclosure.
We have reviewed all AI-assisted work, by manually checking suggested references and for the correctness of AI-generated experimental code. We take responsibility for the final content of this work, including text, claims or artifacts produced with the aid of generative AI.

\subsection*{Ethics statement}

This paper analyzes the representations learned by open-weight models. We do not believe that it introduces any novel ethical hazards beyond those already existing with the models being studied.

\subsection*{Reproducibility statement}

We do not introduce new models, and only analyze existing open-weight models. While we believe that our experimental analysis could be easily reproduced with the information present in the paper, we intend to release the source code for our analysis.



\bibliography{iclr2027_conference}
\bibliographystyle{iclr2027_conference}

\appendix
\section{Appendix}
\subsection{Layer-Wise Results on MIEB}
\label{sec:appendix_mieb}

\begin{figure}[h]
  \begin{minipage}[b]{0.48\linewidth}
    \centering
    \includegraphics[width=\linewidth]{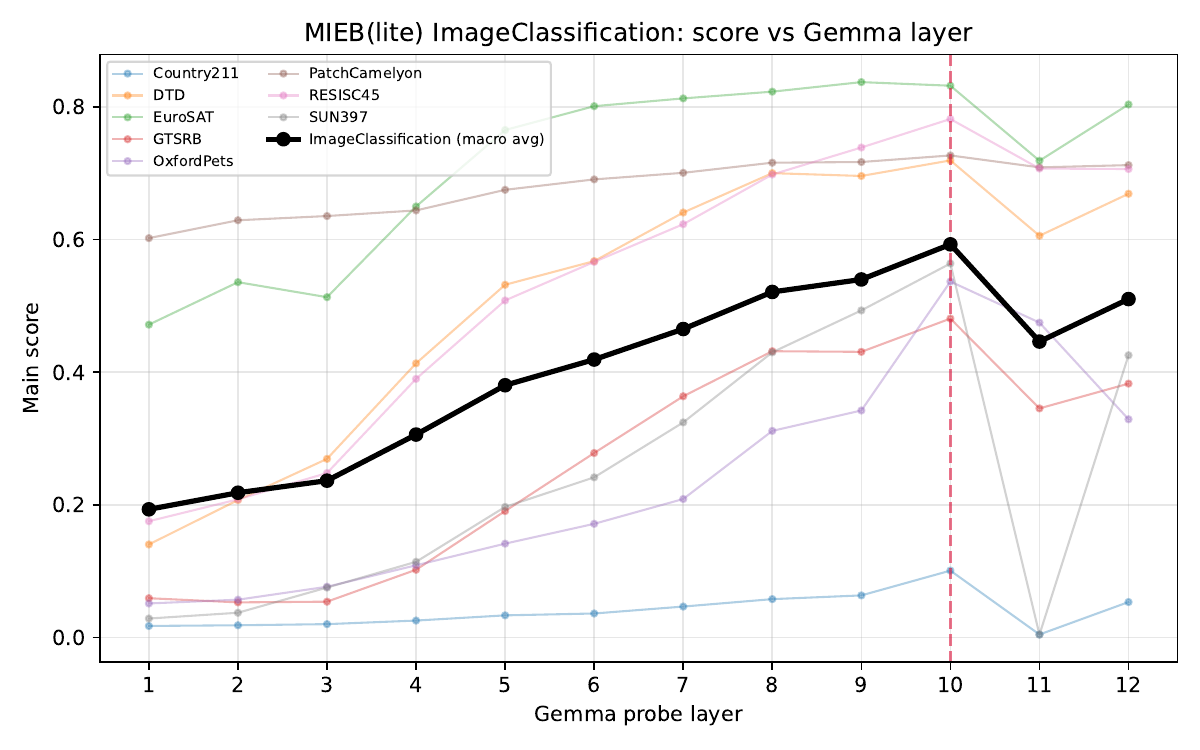}
    \caption{Per-layer sweep for Gemma4-12B on Image Classification Tasks}
    \label{fig:gemma4_image_classification}
  \end{minipage}
  \hspace{0.2cm}
  \begin{minipage}[b]{0.48\linewidth}
    \centering
    \includegraphics[width=\linewidth]{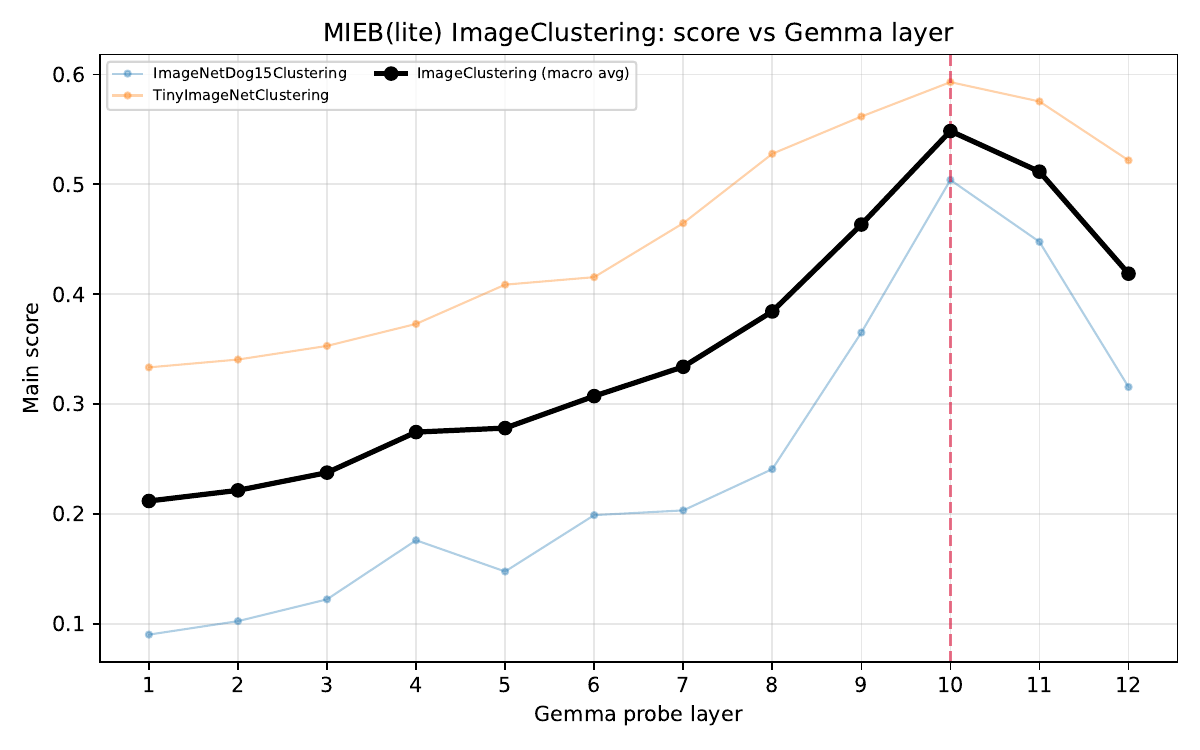}
    \caption{Per-layer sweep for Gemma4-12B on Image Clustering Tasks}
    \label{fig:gemma4_image_clustering}
  \end{minipage}
\end{figure}

\end{document}

%% file: math_commands.tex
\usepackage{amsmath,amsfonts,bm}

\def\eqref#1{equation~\ref{#1}}

\def\1{\bm{1}}

\DeclareMathAlphabet{\mathsfit}{\encodingdefault}{\sfdefault}{m}{sl}
\SetMathAlphabet{\mathsfit}{bold}{\encodingdefault}{\sfdefault}{bx}{n}

